\documentclass[10pt,twocolumn,letterpaper]{article}

\PassOptionsToPackage{pagebackref,breaklinks,colorlinks}{hyperref}
\usepackage{cvpr}              

\usepackage{graphicx}
\usepackage{amsmath}
\usepackage{amssymb}
\usepackage{booktabs}
\usepackage{multirow}
\usepackage{array}
\usepackage{siunitx}
\usepackage{amsfonts}
\usepackage{algorithm}
\usepackage{algpseudocode}
\usepackage{mathtools}
\usepackage{bm}
\usepackage{hyperref}
\usepackage{tikz-cd}
\usepackage[most]{tcolorbox}
\usepackage{dblfloatfix}  
\usepackage{booktabs}   
\usepackage{colortbl}   
\usepackage{xcolor}
\usepackage{tikz}
\usepackage{parskip}
\usepackage{caption}
\usepackage{enumitem}
\setlist[itemize]{noitemsep,topsep=0pt,parsep=0pt,partopsep=0pt}        
\setlist[enumerate]{noitemsep,topsep=0pt,parsep=0pt,partopsep=0pt}      
\setlist[description]{noitemsep,topsep=0pt,parsep=0pt,partopsep=0pt}    
\usepackage{caption}
\newcommand{\best}[1]{\cellcolor{gray!20}\textbf{#1}}
\newcommand*\circled[1]{%
  \tikz[baseline=(char.base), scale=0.5]{  
    \node[
      shape=circle,
      draw,
      inner sep=1pt                
    ] (char) {\tiny #1};           
  }%
}

\usepackage[pagebackref,breaklinks,colorlinks]{hyperref}

\usepackage[capitalize]{cleveref}
\crefname{section}{Sec.}{Secs.}
\Crefname{section}{Section}{Sections}
\Crefname{table}{Table}{Tables}
\crefname{table}{Tab.}{Tabs.}

\def\cvprPaperID{*****} 
\def\confName{CVPR}
\def\confYear{2022}

\begin{document}

\title{Leveraging Distribution Matching to Make Approximate Machine Unlearning Faster}


\author{Junaid Iqbal Khan\\
{\tt\small dianujkotov15@gmail.com}
}

\maketitle

\begin{abstract}
Approximate machine unlearning (AMU) enables models to `forget' specific training data through specialized fine-tuning on a retained (and forget) subset of training set. However, processing this large retained subset still dominates computational runtime, while reductions of unlearning epochs also remain a challenge. 
In this paper, we propose two complementary methods to accelerate arbitrary classification-oriented AMU method. 
First, \textbf{Blend}, a novel distribution-matching dataset condensation (DC), merges visually similar images with shared blend-weights to significantly reduce the retained set size. It operates with minimal pre-processing overhead and is orders of magnitude faster than state-of-the-art DC methods.
Second, our loss-centric method, \textbf{Accelerated-AMU (A-AMU)}, augments the AMU objective to quicken convergence. A-AMU achieves this by combining a steepened primary loss to expedite forgetting with a differentiable regularizer that matches the loss distributions of forgotten and in-distribution unseen data.
Our extensive experiments demonstrate that this dual approach of data and loss-centric optimization dramatically reduces end-to-end unlearning latency across both single and multi-round scenarios, all while preserving model utility and privacy. To our knowledge, this is the first work to systematically tackle unlearning efficiency by jointly designing a specialized dataset condensation technique with a dedicated accelerated loss function.
Code is available at \url{https://github.com/algebraicdianuj/DC_Unlearning}.
\end{abstract}

\begin{figure*}[t]
\begin{center}
\includegraphics[width=0.9\linewidth]{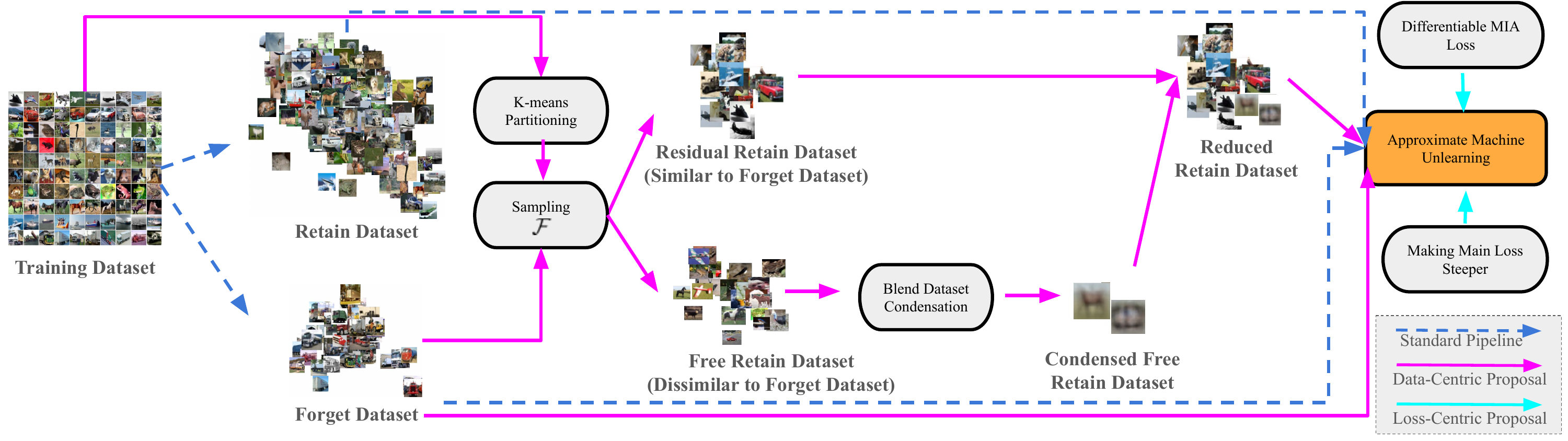}
\end{center}
\caption{The overall pipeline of our proposal to make accelerate AMU, which comprise data-centric and loss-centric components}
\label{fig:short}
\end{figure*}

\vspace{-1ex}
\vspace{-1ex}

\section{Introduction}

Machine unlearning (MU) seeks to expunge specific training points from a deployed model.  \textbf{\emph{Exact MU}} \cite{cao2015towards, bourtoule2021machine} offers perfect deletion guarantees by fully retraining, often via data‐sharding schemes, but remains prohibitively slow for modern networks.  \textbf{\emph{Approximate MU}} (AMU) \cite{jia2023model,kurmanji2023towards,chundawat2023can,golatkar2020eternal,warnecke2021machine,guo2019certified} replaces such guarantees with speed, by achieving following goal in one-step or iterative settings:
\begin{equation}
\min_{\text{model parameters}} \alpha \,\underbrace{\textbf{Main Loss}}_{\text{on retained data}} \,+\, \beta \,\underbrace{\textbf{Forgetting Reg.}}_{\text{drives deletion}}
\tag{*}
\label{eq:template}
\end{equation}

Here, \( \alpha, \beta \ge 0 \) control the utility-vs-forgetting trade-off. We focus on the common iterative case \( \alpha = 1 \), as one step variants assume unrealistically simple loss landscapes. Although approximate machine unlearning (AMU) avoids full retraining, repeatedly scanning the large retained set still dominates runtime. At the same time, reducing unlearning epochs remains a challenge, analogous to how BatchNorm accelerated training convergence~\cite{ioffe2015batch}.

We address this by realizing reduction of the retained dataset via specialized speed-focused dataset condensation \cite{zhao2020dataset}. Building upon distribution-matching dataset condensation \cite{zhao2023dataset}, we propose \textbf{\emph{Blend}}\footnote{This method is a revision of our previous work available at \url{https://arxiv.org/abs/2402.00195}.}, that learns a compact \emph{blend-weight} vector to synthesize a tiny set of synthetic datasets that preserve training utility. Integrating this into AMU pipeline shrinks the retained set and accelerates unlearning without sacrificing accuracy or privacy, with a minor overhead. We further push for loss-centric optimization through Accelerated-AMU (\textbf{\emph{A-AMU}}) loss that preserves the template \eqref{eq:template}, except it steepens the \textbf{main loss} at per sample (or batch) level, and regularizes it with a distribution matching formulation of a membership inference attack (MIA) minimizer. This enables rapid convergence to the unlearned model. Our approach demonstrates substantial speedups and very strong unlearning compared to baseline SOTA AMU methods. We summarize our contributions as follows:
\begin{itemize}
\item \textbf{\emph{Blend}} DC, specifically designed for unlearning, combines image blending with distribution matching, achieving an average 44\% retained dataset reduction with approximately 7.5\% preprocessing overhead compared to baseline unlearning, and is approximately 1500 times faster than state-of-the-art dataset condensation methods.
\item \textbf{\emph{A-AMU}} augments the AMU loss formulation with a steepened primary loss and a differentiable membership-inference regularizer, reducing unlearning runtime by 76.82\% in single-round and 51.83\% in multi-round scenarios compared to baseline methods, by minimizing the number of epochs, while maintaining strong privacy-utility statistics.
\item Our benchmarks show 84.61\% and 54.45\% faster end-to-end unlearning in single and multi-round scenarios, respectively, on various image classification datasets and models, effectively filling latency gaps compared to current state-of-the-art AMU methods.
\end{itemize}
\vspace{-1ex}
\vspace{-1ex}

\section{Related Works}

Machine unlearning (MU) has evolved along two complementary axes. \textbf{\emph{Exact MU}} guarantees equivalence to full retraining by either retraining on data shards (SISA \cite{bourtoule2021machine}, ARCANE \cite{yan2022arcane}) or deriving closed-form updates for simple models such as $k$-means and linear classifiers \cite{ginart2019making,mahadevan2021certifiable}, but scaling these ideas to over-parameterized deep networks remains hard. \textbf{\emph{Approximate MU}} relaxes this guarantee for speed: influence-function or Hessian sketches explicitly subtract a point’s contribution \cite{graves2021amnesiac,wu2020deltagrad,golatkar2020eternal,warnecke2021machine};``bad-teacher" distillation removes the target data then distills utility back from a clean model \cite{kurmanji2023towards,chundawat2023can}; and recent accelerations prune parameters \cite{tanaka2020pruning,jia2023model}, nudge decision boundaries \cite{chen2023boundary}, or couple sparse adapters with shard partitioning, culminating in class-agnostic, data-free DELETE \cite{zhou2025decoupled} and Unlearn-and-Distill \cite{lee2025distillation}. Parallel to MU, \textbf{\emph{dataset condensation (DC)}} compresses training sets into compact proxies: from gradient matching \cite{wang2018dataset,zhao2020dataset} and distribution/trajectory matching \cite{zhao2023dataset,cazenavette2022dataset} to latent re-parameterization, differentiable augmentation and infinite-width kernels \cite{kim2022dataset,zhao2021dataset,nguyen2021dataset}, with EDF recently highlighting class-discriminative cues via Grad-CAM \cite{wang2025emphasizing}. However \textbf{\emph{DC for MU}} is nascent: QuickDrop naively integrates gradient-matching DC into federated unlearning \cite{dhasade2024quickdrop}, and TCGU shows graph-level gains by naively condensing once then editing the proxy \cite{li2024tcgu}; nevertheless, existing DC objectives are still too costly to serve as a drop-in unlearning accelerator, motivating our targetted and simpler, distribution-matching approach.

\vspace{-1ex}
\vspace{-1ex}

\section{Methodology}
\subsection{Preliminaries}
Let the training dataset $\mathbf{D}$, consist of labeled images, denoted as \(\{I_p\}_p\), comprising of \(C\) classes and \(N_c\) images per class, where \(c \in \{1, 2, \dots, C\}\). 
The training dataset is divided into retain $\mathbf{R}$ and forget $\mathbf{F}$ dataset. $\mathbf{D}=\{I_p\}_p$ was used to train a parameterized target network, and it achieved local minima parameters \(\theta\) on average loss \(\mathcal{L}(\theta; x, y)\) over \((x, y) \in \mathbf{R} \cup \mathbf{F}\). Let \( \mathbf{D} \) and \( \mathbf{\tilde{T}} \) be datasets drawn from the same underlying distribution \( \mathcal{D} \), where \( \mathbf{\tilde{T}} \) is the testing dataset, and is used in evaluation of training. Our objective is to unlearn forget set \(\mathbf{F} \subset \{I_p\}_p\) from the pretrained model. Denote by \(\mathcal{D}_R\) and \(\mathcal{D}_F\) the empirical distribution of \(\mathbf{R}\) and \(\mathbf{F}\) respectively. The subset $\mathbf{T} \subseteq \mathbf{\tilde{T}}$ consists of elements whose class labels match those in \( \mathbf{F} \), and \(\mathcal{D}_T\) is its empirical distribution.

\subsection{Blend Dataset Condensation}

Using \textbf{Algorithm~1}, we divide images of each class of $\textbf{D}$ into $k$ sub-classes using $k$-means clustering over the feature representations, extracted by a randomly initialized lightweight feature extractor $\psi_{\vartheta}$ (much smaller than the pretrained model to be unlearned). The resulting partition of the dataset can be written as:
\begin{equation}
\begin{split}
\{I_p\}_p \xrightarrow{\text{k-means partitioning}} \left\{ \left\{ I_{ij} \right\}_{j=1}^{n_i} \right\}_{i=1}^{k \times C}
\end{split}
\end{equation}
where $i$ indexes the sub-class (with a total of $k \times C$ sub-classes), and $j$ indexes the images within each sub-class. Each $i$-th subclass contain $n_i$ images.
For each sub-class $i$, we define a \emph{blended image} $\mathcal{I}(\omega_i)$ as a weighted average of the images $\{ I_{ij} \}_{j=1}^{n_i}$ b   elonging to that sub-class:
\vspace{-1ex}
\vspace{-1ex}
\begin{equation} \label{eq:importance_weighted_average}
\mathcal{I}(\omega_i)
\overset{\mathrm{blend}}{=}
\frac{1}{\sum_{j=1}^{n_i} \omega_{ij}}
\sum_{j=1}^{n_i} \omega_{ij}\,I_{ij}
\end{equation}
\vspace{-1ex}
\vspace{-1ex}

\begin{algorithm}[t]
\caption{$k$-means Partitioning ($\mathbf{D}=\{I_p\}_p$, $\psi_{\vartheta}$)}
\label{alg:kmeans_partition}
\begin{footnotesize}
\begin{algorithmic}[1]
  \Statex \textbf{Indices:} $c = 1,\dots,C$ (class); $\kappa = 1,\dots,k$ ($k$-means cluster); $b = 1,\dots,n_\kappa$ (images in cluster $\kappa$)
  \State $\mathcal{C} \gets \varnothing$ \Comment{Initialize image clusters}
  \For{each class $c = 1,\dots,C$}
    \State $\{I_a\}_{a=1}^{N_c} \gets \{I_p \mid y_p = c\}$ \Comment{Collect class images}
    \State $\{\psi_{\vartheta}(I_a)\}_{a=1}^{N_c}$ \Comment{Compute feature vectors}
    \State $\{\psi_{\vartheta}(I_a)\}_{a=1}^{N_c} \xrightarrow{k\text{-means}} \bigl\{\{\psi_{\vartheta}(I_{b\kappa})\}_{b=1}^{n_\kappa}\bigr\}_{\kappa=1}^{k}$ \Comment{Run $k$-means}
    \State $\bigl\{\{\psi_{\vartheta}(I_{b\kappa})\}_{b=1}^{n_\kappa}\bigr\}_{\kappa=1}^{k} \to \bigl\{\{I_{b\kappa}\}_{b=1}^{n_\kappa}\bigr\}_{\kappa=1}^{k}$ \Comment{Map to images}
    \State $\mathcal{C} \gets \mathcal{C} \cup \bigl\{\{I_{b\kappa}\}_{b=1}^{n_\kappa}\bigr\}_{\kappa=1}^{k}$ \Comment{Add to global set}
  \EndFor
  \State $\mathcal{C} = \bigl\{\{\{I_{b\kappa c}\}_{b=1}^{n_{\kappa c}}\}_{\kappa=1}^{k}\bigr\}_{c=1}^{C}$ \Comment{Full cluster set}
  \State \textbf{Refine notation:} $b \mapsto i$, $(\kappa,c) \mapsto j$
  \State \Return $\{\{I_{ij}\}_{i=1}^{n_j}\}_{j=1}^{k \times C}$ \Comment{Partitioned \textbf{D} dataset}
\end{algorithmic}
\end{footnotesize}
\end{algorithm}

\vspace{-1ex}
\vspace{-1ex}
where $\omega_{ij}$ are scalar weights associated with the $j$-th image in sub-class $i$. These weights are treated as learnable parameters and are optimized to ensure that the feature representation of the blended image $\mathcal{I}_i$ closely matches the mean feature representation of its constituent images to form a condensed single image proxy of $\{ I_{ij} \}_{j=1}^{n_i}$. We estimate $\omega_i$ in \eqref{eq:importance_weighted_average}, via following distribution matching \cite{zhao2023dataset} loss:

\begin{tcolorbox}[
    colframe=black,
    colback=white,
    sharp corners,
    enhanced,
    boxsep=1pt,
    left=1pt,
    right=1pt,
    top=1pt,
    bottom=1pt,
    fontupper=\footnotesize,  
    halign=center             
]
\(\displaystyle              
  \min_{\omega_i}\;
  \mathbb{E}_{\vartheta \sim P_{\vartheta}}
  \Bigl\|
    \tfrac{1}{n_i}\sum_{j=1}^{n_i}\psi_{\vartheta}(I_{ij})
    \;-\;
    \psi_{\vartheta}\bigl(\mathcal{I}(\omega_i)\bigr)
  \Bigr\|_2^2
\)
\end{tcolorbox}

where $\psi_{\vartheta}$ is the feature extractor parameterized by $\vartheta$, and $\|\cdot\|$ denotes the Euclidean norm. The expectation is taken over a distribution $P_{\vartheta}$ of lightweight feature extractors to promote robustness, as adapted in \cite{zhao2023dataset}.

\subsection{Creating the \emph{Reduced} Retain Dataset}

Dataset condensation replaces each original feature
$x\!\in\!\mathbb{R}^{n}$ (as Lipchitz mapping of original image) with a synthetic prototype whose placement
preserves the topology of the manifold on which the features live.
Because this process is global, a prototype can still migrate into the
neighbourhood of the to-be forgotten feature samples, thus creating an obstacle for unlearning. Furthermore, since the condensation time scales with the size of the retain set, condensing the entire retain dataset becomes a significant computational bottleneck.
We therefore propose a targeted dataset condensation on specific partition of retain dataset, whose features are most far away from those of forget samples.  To achieve this, we propose following steps:
\vspace{-1ex}
\vspace{-1ex}
\paragraph{Step 1: Partition the retain set with sampling $\mathcal{F}$}
Using \textbf{Algorithm~2}, we systematically separate the retain images (using partitioned training set and $\mathbf{F}$) into:
\begin{itemize}
  \item \emph{free images} $\mathbf{R}_{\text{free}}$ whose features are
        \textbf{distant} from~$\mathbf{F}$
  \item \emph{residual images} $\mathbf{R}_{\text{residual}}$ that lie
        \textbf{close} to features of ~$\mathbf{F}$
\end{itemize}
Specifically, the inter-cluster indicator $\phi(j)$ and the intra-cluster indicator $\Phi(i)$ test whether the set of images 
\(\{I_{ij}\}_{j=1}^{n_i}\) for each cluster \(j\) intersects the forget set \(\mathbf{F}\), thereby partitioning $\mathbf{R}$ into $\mathbf{R}_{\text{free}}$ and $\mathbf{R}_{\text{residual}}$.

\vspace{-1ex}
\vspace{-1ex}
\paragraph{Step 2: Condense only the free retain subset}
Each cluster in the free retain subset (i.e.\ those with \(\phi(j)\neq -1\)) is condensed by blending its \(n_j\) images into a single prototype, thus creating condensed version of free retain set:
\begin{equation}\label{eq:distribution_matching_loss}
\mathbf{Condense}\bigl(\mathbf{R}_{\text{free}}\bigr)
\;=\;
\Bigl\{
  \{\,I_{ij}\,\}_{j=1}^{n_i}
  \;\xrightarrow{\;\mathrm{blend}\;}
  \mathcal{I}\!\bigl(\omega_i\bigr)
\Bigr\}_{i=\phi(1)}^{\phi(kC)} .
\end{equation}
allowing no synthetic point
converging towards ~$\mathbf{F}$ features.

\paragraph{Step 3: Create the \emph{Reduced} Retain Dataset}
Having partitioned the original retain set into the distant “free” clusters and the forget images sensitive “residual” images, and condensed only the former via the blending operation in Eq.~\eqref{eq:distribution_matching_loss}, we now assemble our final training set.  
\textbf{\[
  \mathbf{R}_{\text{reduced}}
  \;=\;
  \underbrace{\mathbf{Condense}\bigl(\mathbf{R}_{\text{free}}\bigr)}
    _{\text{condensed,\;low-importance}}
  \;\cup\;
  \underbrace{\mathbf{R}_{\text{residual}}}
    _{\text{uncondensed,\;high-importance}}.
\]}%
This reduced retain dataset preserves all samples that lie near the forget manifold (ensuring precise boundary control during unlearning), while summarising only those clusters that have no intersection with \(\mathbf{F}\).  As a result, $\mathbf{R}_{\text{reduced}}$ offers the computational savings from total dataset condensation, with advantage of not sacrificing the fine‐grained feature information needed to effectively forget and persist generalization, which would have been possible with raw AMU.

\begin{algorithm}[t]
\caption{Sampling $\mathcal{F} \bigl (\mathbf{F}, \{\{I_{ij}\}_{i=1}^{n_j}\}_{j=1}^{k\times C}\bigr)$}
\label{alg:sampling}
\begin{footnotesize}
\begin{algorithmic}[1]
  \Statex \textbf{Indices:} $j = 1,\dots,kC$ (cluster); $i = 1,\dots,n_j$ (image in cluster $j$)
  \State $\phi(j) = \begin{cases}
      -1, & \mathbf{F}\cap\{I_{ij}\}_{j=1}^{k\times C}=\emptyset \\
      j, & \text{otherwise}
    \end{cases}$
    \Comment{indexing clusters with no $\mathbf{F}$}
  \State $\Phi(i) = \begin{cases}
      i, & \mathbf{F}\cap\{I_{ij}\}_{i=1}^{n_j} \\
      -1, & \text{otherwise}
    \end{cases}$
    \Comment{indexing within $\mathbf{F}$ clusters}
  \State $\mathbf{R}_{\text{free}} = \{\{I_{ij}\}_{i=1}^{n_j}\}_{j=\phi(1)}^{\phi(kC)}$
    \Comment{$\mathbf{R}$ subset dissimilar to $\mathbf{F}$}
  \State $\mathbf{R}_{\text{residual}} = \{\{I_{ij}\}_{i=\Phi_j(1)}^{\Phi_j(n_j)}\}_{j=1}^{kC}$
    \Comment{$\mathbf{R}$ subset similar to $\mathbf{F}$}
  \State \Return $\mathbf{R}_{\text{free}}, \mathbf{R}_{\text{residual}}$
\end{algorithmic}
\end{footnotesize}
\end{algorithm}

\subsection{Accelerated Approximate Unlearning}
Carlini \emph{et al.}~\cite{carlini2022membership} report that per-sample classification losses of deep networks are \emph{strongly non-Gaussian}.
They considered to “Gaussianise’’
losses by ad-hoc monotone maps (e.g. combination of $\exp$ and $\log$ maps), without
guaranteeing that the output is normally distributed.
We instead apply the probability–integral transform (PIT), i.e. for any continuous variable \(X\) with CDF \(F_X\),
\(Z=\Phi^{-1}\!\bigl(F_X(X)\bigr)\) is standard normal, where \(\mathbf{\Phi^{-1}}\) is the probit (quantile) function of the standard
normal distribution \(\mathcal N(0,1)\) (mean \(0\), variance \(1\)).
Substituting \(F_X\) by a differentiable empirical surrogate yields an
end-to-end differentiable pipeline whose transformed losses converge to
\(\mathcal N(0,1)\).
\\
Following this motivation, we intend to transform arbitrary random samples, which may not be following Gaussian distribution towards samples following standard normal distribution. Let $
x=(x_1,\ldots,x_N)
\;\stackrel{\text{i.i.d.}}{\sim}\;F$, such that 
$F$ is an unknown continuous CDF.
The empirical CDF at \(x_i\) is
\(
\hat F_N(x_i)=\tfrac1N\sum_{j=1}^{N}\mathbf 1\{x_j\le x_i\}.
\)
Replacing the indicator by the temperature–controlled sigmoid  
\(
\sigma_K(z)=(1+e^{-Kz})^{-1},\;K>0,
\)
gives the differentiable estimator
\[
\begin{aligned}
Q_i^{(K)}(x) &= \frac1N\sum_{j=1}^{N}\sigma_K\!\bigl(x_i-x_j\bigr),
\qquad i=1,\ldots,N,\\[2pt]
\end{aligned}
\]
such that $\lim_{K\to\infty} Q_i^{(K)}(x) = \hat{F}_N(x)$. We stack all samples into coordinates of vector as

\begin{equation}\label{eq:quantiles}
\mathbf Q^{(K)}(x)
=\bigl(Q_1^{(K)}(x),\ldots,Q_N^{(K)}(x)\bigr)^{\!\top}\in(0,1)^N
\end{equation}

Because each entry of \(\mathbf Q^{(K)}(x)\) lies in \((0,1)\),
apply the probit function to obtain \cite{van2007transformation}

\begin{equation}\label{eq:transformation}
\mathbf z^{(K)}=\mathbf{\Phi^{-1}}\!\bigl(\mathbf Q^{(K)}(x)\bigr)\in\mathbb R^{N}.
\end{equation}

Thus any sample $x$ from a continuous distribution is converted to a
sample whose coordinates converge to i.i.d.\ standard normals as
\(K\to\infty\): $
\mathbf z^{(K)} \;\xrightarrow{K\to\infty}\;
\mathbf{\Phi^{-1}}\!\bigl(\hat F_N(x)\bigr).
$
A finite but large $K$ balances smooth
gradients (for optimization) with true empirical
CDF proximity.

\subsubsection{Transforming losses of model to standard normal}

For the frozen pretrained network with parameters $\theta$, we define the per-sample log transformed network's original loss
(e.g.\ cross-entropy) as $
\ell(x,y)=\log\!\bigl(1+\mathcal{L}(\theta;x,y)\bigr)$.

Collect all losses of $\mathbf F$ and $\mathbf T$:
\[
\begin{aligned}
\boldsymbol{\ell}_F &=
\bigl(\ell(x_1,y_1),\dots,\ell(x_{n_F},y_{n_F})\bigr)^{\!\top},\\[2pt]
\boldsymbol{\ell}_T &=
\bigl(\ell(x'_1,y'_1),\dots,\ell(x'_{n_T},y'_{n_T})\bigr)^{\!\top}.
\end{aligned}
\]

with $(x_i,y_i)\!\sim\!\mathcal{D}_{F}$ and
$(x'_i,y'_i)\!\sim\!\mathcal{D}_{T}$.

Applying the vector transformation from \eqref{eq:transformation} to the
loss vectors yields $
\mathbf Z_F^{(K)}=\mathbf{\Phi^{-1}}\!\bigl(\mathbf Q^{(K)}(\boldsymbol{\ell}_F)\bigr),
\qquad
\mathbf Z_T^{(K)}=\mathbf{\Phi^{-1}}\!\bigl(\mathbf Q^{(K)}(\boldsymbol{\ell}_T)\bigr),
$
where $\mathbf Q^{(K)}$ is the differentiable CDF estimator from \eqref{eq:quantiles}.
As $K\to\infty$ the mapping converges to the empirical probit, and this allows use to play with transformed losses in distribution matching manner. As emphasize previously, 
practice we keep $K$ sufficiently large to preserve
differentiability.

\subsubsection{Membership-Inference Attack (MIA) Loss}

With the Gaussian kernel
\(
\curlywedge(z,z')=\exp\!\bigl(-\|z-z'\|^{2}/2\bigr),
\)
the squared maximum mean discrepancy between the transformed loss
distributions is
\begin{equation}\label{eq:mmd_loss}
\begin{aligned}
\mathcal{L}_{\mathrm{MMD}}(\theta)
&=
\mathbb{E}_{z,z'\sim\mathbf Z_F^{(K)}}\!\bigl[\curlywedge(z,z')\bigr]
+
\mathbb{E}_{\tilde z,\tilde z'\sim\mathbf Z_T^{(K)}}\!\bigl[\curlywedge(\tilde z,\tilde z')\bigr]\\
&\quad
-\,2\,
\mathbb{E}_{z\sim\mathbf Z_F^{(K)},\,\tilde z\sim\mathbf Z_T^{(K}}\!\bigl[\curlywedge(z,\tilde z)\bigr].
\end{aligned}
\end{equation}
Minimising $\mathcal{L}_{\mathrm{MMD}}$ aligns the transformed loss
distribution of the forget set with that of the class-matched test
subset, thereby mitigating membership-inference leakage while remaining
fully differentiable in $\theta$ and $K$.

\vspace{-1ex}
\vspace{-1ex}

\subsubsection{Accelerated (A-) AMU Objective}

We come to the second thesis of this paper, i.e. to accelerate unlearning at unlearning objective level. We propose following $\textbf{A-AMU}$ minimizing objective, which proposes \textbf{\emph{two main solutions}} over a general machine unlearning loss objective (mentioned in `Introduction' section).

\begin{tcolorbox}[
    colframe=black,
    colback=white,
    sharp corners,
    enhanced,
    boxsep=1pt,     
    left=1pt,
    right=1pt,
    top=1pt,
    bottom=1pt,
    fontupper=\footnotesize,   
    halign=center
]
\(\displaystyle
  \min_{\theta}\,
    \underbrace{\mathbb{E}_{(x,y)\sim \mathcal{D}_R}\!\bigl[\mathcal{L}(\theta;x,y)^2\bigr]}_{\textbf{Steeper Main Loss}}
    \;+\;
    \lambda\,\underbrace{\mathcal{L}_{\mathrm{MMD}}(\theta)}_{\textbf{Diff. MIA}}
    \;+\;
    \underbrace{\Gamma(\theta;\mathbf{R},\mathbf{F})}_{\textbf{Other Reg.}}
\)
\end{tcolorbox}

where $\lambda\!>\!0$ is regularization parameter.

\paragraph{Solution \# 1: Steeper main loss.}
Replacing per-sample \(\mathcal{L}\) by its square
\(\mathcal{L}^2\) multiplies the per-sample gradient magnitude by \(2\,\mathcal{L}\).
Consequently, gradient descent updates make larger
progress early in training, in such a way that large errors on~$\mathbf R$ are penalized disproportionately, while small errors become less important in course of optimization. So the model departs more rapidly from the decision boundaries, which it learned earlier during training schedule. In order to prevent overfitting on high error retain dataset samples, we take square of average loss per batch, instead of per sample squaring. This also induces stability on choice of learning rate. Infact, in all of our experiments, we set a fixed learning rate for our approach on all models and datasets (including other hyperparameters), which is in strong contrast from other SOTA \textbf{AMU} methods that require hectic finetuning on each setting.

\vspace{-1ex}
\vspace{-1ex}\vspace{-1ex}

\paragraph{Solution \# 2: Differentiable MIA term.}
$\mathcal{L}_{\mathrm{MMD}}(\theta)$, from \eqref{eq:mmd_loss}, essentially models something absolute in unlearning, i.e. for unlearned model, forget samples appears the same as test samples, which supersedes subjective modelling of forgetting regularizations in \cite{kurmanji2023towards,jia2023model, chundawat2023can} dependent upon accuracy or parameter distribution hypothesis of unlearning.
Minimizing this term over $\theta$ suppresses membership-inference
signals on~$\mathbf F$ while remaining fully differentiable in~$\theta$. This interplays with making main loss steeper, as the parameter rapidly shift to overfit on retain set (or its reduced version), the differentiable MIA term allow quick restriction of parameters to only those trajectory, where MIA score is minimum.
\vspace{-1ex}
\vspace{-1ex}
\vspace{-1ex}
\paragraph{Steeper main loss + Differentiable MIA term.}
Combining the controlled instability from steeper main loss, around decision boundaries, with reduction of training trajectories towards absolute unlearning solutions, forms a strong acceleration of unlearning. As would be seen later in our experiments,
$\textbf{A-AMU}$ objective minimization achieves competitive MIA suppression
and retain accuracy after $
\sim 1 \, \text{epoch}$. 

\vspace{-1ex}
\vspace{-1ex}
\vspace{-1ex}

\paragraph{Additional Regularisation \(\Gamma\).}
The regularization term \(\Gamma(\theta; \mathbf{R}, \mathbf{F})\) serves as a characteristic placeholder for AMU unlearning ideas like weight sparsity \cite{jia2023model}, negative distillation \cite{kurmanji2023towards}, bad teacher distillation \cite{chundawat2023can}, etc.

\section{Experiments}
\subsection{Experimental Setup}

\paragraph{Methodology}
We evaluate our methods on CIFAR-10, SVHN, and CINIC-10 datasets using ResNet-18 and ResNet-50 models. Our approach is benchmarked against six AMU methods: \textbf{Retraining}, \textbf{Catastrophic Forgetting (CF)} (fine-tuning) \cite{kurmanji2023towards, jia2023model}, \textbf{Bad Distillation} \cite{chundawat2023can}, $\mathbf{L_1}$-Sparsity \cite{jia2023model}, \textbf{Pruning} \cite{jia2023model}, and \textbf{SCRUB} \cite{kurmanji2023towards}. For a fair comparison, implementations were adapted from their official repositories, and hyperparameters were rigorously fine-tuned for each model and dataset.\footnote{Complete hyperparameter configurations are available in the 'hyperparameters' folder of our code repository: \url{https://github.com/algebraicdianuj/DC_Unlearning}}.
We categorize these baselines into three tiers for analysis: \emph{naive} (\textbf{Retraining}, \textbf{CF}), \emph{non-naive} ($\mathbf{L_1}$-sparse, \textbf{Pruning}, \textbf{SCRUB}, \textbf{Bad Distillation}), and our proposed \emph{accelerated} (\textbf{A-}) methods.

\vspace{-1ex}
\vspace{-1ex}

{
\scriptsize                     
\renewcommand{\arraystretch}{0.8}   
\setlength{\tabcolsep}{4pt}         
\begin{table*}[t]
\centering
\caption{Unlearning on ResNet-50/CIFAR-10 (mean $\pm$ 95\% CI) with $k$-means partitioning time $5.5753 \pm 0.1391$ s, $\mathcal{F}$-sampling time $0.1184 \pm 0.0107$ s}
\label{tab:resnet50_32_cifar100}
\resizebox{\textwidth}{!}{
\begin{tabular}{lcccccccccc}
\toprule
\multicolumn{1}{c}{} & 
\multicolumn{2}{c}{\textbf{MIA Score~$\downarrow$}} & 
\multicolumn{2}{c}{\textbf{Retain Accuracy~$\uparrow$}} & 
\multicolumn{2}{c}{\textbf{Forget Accuracy~$\downarrow$}} & 
\multicolumn{2}{c}{\textbf{Test Accuracy~$\uparrow$}} & 
\multicolumn{2}{c}{\textbf{Unlearning Time (s)~$\downarrow$}} \\
\cmidrule(lr){2-3} \cmidrule(lr){4-5} \cmidrule(lr){6-7} \cmidrule(lr){8-9} \cmidrule(lr){10-11}
\textbf{Method} & \textbf{w/o DC} & \textbf{w DC} & \textbf{w/o DC} & \textbf{w DC} & \textbf{w/o DC} & \textbf{w DC} & \textbf{w/o DC} & \textbf{w DC} & \textbf{w/o DC} & \textbf{w DC} \\
\midrule
\multicolumn{11}{c}{\textbf{Random Class Forgetting (Condensation Time for 1000 IPC: $47.7812 \pm 1.3323$ s leading to 80.0\% retain dataset reduction)}} \\
\midrule
Retraining       & \best{$48.61 \pm 1.85$} & \best{$49.49 \pm 2.38$} & \best{$100.00 \pm 0.00$} & $68.45 \pm 2.51$ & \best{$0.00 \pm 0.00$} & \best{$0.00 \pm 0.00$} & $79.96 \pm 1.10$ & $56.46 \pm 2.81$ & $3580.71 \pm 94.46$ & $705.82 \pm 0.30$ \\
CF               & $51.07 \pm 3.71$ & $57.87 \pm 2.74$ & $100.00 \pm 0.00$ & $97.89 \pm 0.18$ & $0.00 \pm 0.00$ & $1.65 \pm 1.73$ & \best{$81.28 \pm 1.33$} & $79.11 \pm 1.89$ & $1789.40 \pm 47.15$ & $352.38 \pm 0.74$ \\
Bad Distillation & $57.97 \pm 5.13$ & $62.99 \pm 6.81$ & $97.31 \pm 1.59$ & \best{$99.69 \pm 0.09$} & $0.00 \pm 0.00$ & $0.01 \pm 0.03$ & $77.68 \pm 2.36$ & \best{$80.64 \pm 1.16$} & $1370.90 \pm 34.32$ & $572.67 \pm 0.69$ \\
$L_1$ Sparse     & $49.18 \pm 2.20$ & $58.95 \pm 4.72$ & $98.52 \pm 0.55$ & $98.54 \pm 0.32$ & $0.00 \pm 0.00$ & $4.03 \pm 3.58$ & $77.61 \pm 2.12$ & $79.61 \pm 1.49$ & $1446.43 \pm 6.34$ & $288.65 \pm 0.31$ \\
SCRUB            & $50.00 \pm 0.00$ & $50.00 \pm 0.00$ & $15.50 \pm 3.00$ & $11.14 \pm 0.06$ & $0.00 \pm 0.00$ & $0.00 \pm 0.00$ & $13.88 \pm 2.72$ & $10.05 \pm 0.09$ & $759.90 \pm 9.19$ & $198.01 \pm 0.21$ \\
Pruning          & $48.37 \pm 2.43$ & $50.56 \pm 0.81$ & $97.10 \pm 1.14$ & $99.08 \pm 0.63$ & $0.00 \pm 0.00$ & $0.00 \pm 0.00$ & $76.85 \pm 1.84$ & $69.80 \pm 3.61$ & $722.06 \pm 5.91$ & $143.73 \pm 0.23$ \\
A-CF  & $52.10 \pm 1.50$ &
$50.99 \pm 0.67$ &
$97.75 \pm 1.50$ &
$95.90 \pm 2.73$ &
$0.00 \pm 0.00$ &
$0.10 \pm 0.19$ &
$77.03 \pm 1.08$ &
$75.90 \pm 3.82$ &
\best{$218.30 \pm 0.26$} &
\best{$44.54 \pm 1.80$}\\

\midrule
\multicolumn{11}{c}{\textbf{10\% Uniform Forgetting (Condensation Time for 1000 IPC: $34.1780 \pm 0.7403$ s leading to 39.8\% retain dataset reduction)}} \\
\midrule
Retraining       & \best{$49.46 \pm 0.87$} & \best{$49.46 \pm 1.29$} & \best{$100.00 \pm 0.00$} & $91.26 \pm 0.59$ & $87.41 \pm 1.05$ & $80.88 \pm 0.78$ & $87.01 \pm 0.16$ & $80.41 \pm 0.64$ & $3554.87 \pm 92.12$ & $2095.34 \pm 29.30$ \\
CF               & $53.32 \pm 0.90$ & $54.20 \pm 0.56$ & $100.00 \pm 0.00$ & \best{$97.79 \pm 0.22$} & $94.34 \pm 0.89$ & $95.19 \pm 0.36$ & \best{$88.64 \pm 0.15$} & \best{$87.52 \pm 0.83$} & $1770.13 \pm 13.17$ & $1046.32 \pm 16.59$ \\
Bad Distillation & $84.07 \pm 1.69$ & $88.42 \pm 1.62$ & $94.77 \pm 0.30$ & $96.16 \pm 0.92$ & \best{$23.76 \pm 9.60$} & \best{$11.93 \pm 1.87$} & $79.80 \pm 1.24$ & $81.67 \pm 1.86$ & $1371.63 \pm 35.81$ & $1297.61 \pm 15.55$ \\
$L_1$ Sparse     & $52.10 \pm 1.60$ & $53.25 \pm 0.38$ & $98.25 \pm 0.39$ & $95.31 \pm 0.91$ & $87.41 \pm 1.22$ & $90.13 \pm 1.93$ & $84.25 \pm 1.17$ & $84.07 \pm 0.90$ & $1463.85 \pm 38.41$ & $856.47 \pm 13.84$ \\
SCRUB            & $51.97 \pm 0.91$ & $52.32 \pm 2.80$ & $97.99 \pm 3.40$ & $86.19 \pm 33.28$ & $87.38 \pm 6.03$ & $83.37 \pm 30.90$ & $84.77 \pm 3.59$ & $78.17 \pm 27.14$ & $756.68 \pm 1.36$ & $483.13 \pm 5.28$ \\
Pruning          & $51.76 \pm 0.54$ & $51.43 \pm 0.93$ & $96.81 \pm 1.42$ & $97.62 \pm 0.39$ & $86.66 \pm 0.69$ & $84.71 \pm 0.77$ & $83.40 \pm 1.58$ & $81.61 \pm 0.68$ & $730.16 \pm 20.13$ & $426.54 \pm 6.89$ \\
A-CF  & $50.20 \pm 2.65$ &
$52.27 \pm 1.46$ &
$95.49 \pm 4.84$ &
$97.06 \pm 2.01$ &
$82.57 \pm 3.42$ &
$84.83 \pm 2.89$ &
$81.72 \pm 5.24$ &
$84.03 \pm 1.15$ &
\best{$218.79 \pm 0.34$} &
\best{$133.32 \pm 5.23$}
\\
\bottomrule
\end{tabular}}
\end{table*}
}
\paragraph{Evaluation}
Unlearning performance is assessed across five key metrics: \textbf{Test Accuracy} ($\uparrow$), \textbf{Retain Accuracy} ($\uparrow$), \textbf{Forget Accuracy} ($\downarrow$), \textbf{Unlearning Time} ($\downarrow$), and a \textbf{MIA Score} ($\downarrow$) for privacy. The MIA score is derived from a powerful white-box LiRA attack \cite{carlini2022membership}, providing a robust measure of privacy risk. For our \textbf{Blend} method, the $k$-means partitioning feature extractor is a lightweight CNN, $35\times$ and $74\times$ smaller than ResNet-18 and ResNet-50 respectively, ensuring its preprocessing overhead remains minimal. All experiments were conducted on a single NVIDIA RTX 3070 GPU (8GB VRAM), except for experiments in Sections 4.4 and 4.5, which were performed on a laptop RTX 3070 (8GB VRAM) due to computational resource availability during the extended experimental phase.

\vspace{-1ex}
\vspace{-1ex}

\begin{figure}[tb]
\centering
\includegraphics[height=1.0\linewidth, width=1.0\linewidth]{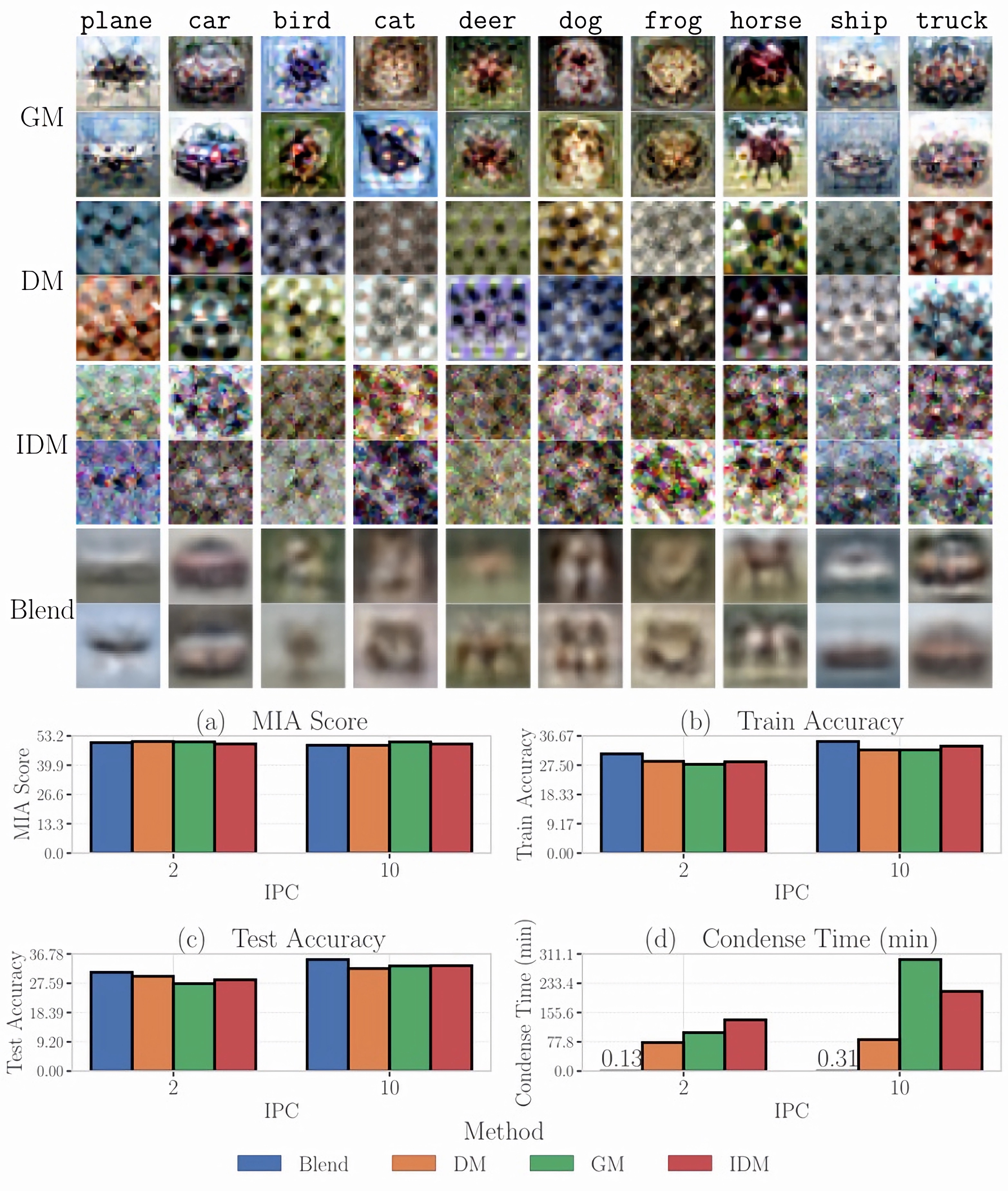}
\caption{Qualitative and quantitative comparison of condensation methods on CIFAR-10/CNN.
Top: class-wise synthetic prototypes generated by \textbf{GM}, \textbf{DM}, \textbf{IDM}, and our \textbf{Blend} method. Bottom: Performance metrics for each method over course of IPC = 2, 10.}
\label{fig:short}
\end{figure}

\vspace{-1ex}

\paragraph{Categorization: }We categorize baselines into three tiers: \emph{naive} (\textbf{Retraining}, \textbf{CF}), \emph{non-naive} ($\mathbf{L_1}$-sparse, \textbf{Pruning}, \textbf{SCRUB}, \textbf{Bad Distillation}), and \emph{accelerated} (\textbf{A-}), the last of which can contain acceleration of all naive and non-naive AMU methods.

\vspace{-1ex}
\vspace{-1ex}

\subsection{The Efficiency and Efficacy of Blend Condensation}
Our proposed dataset condensation method, \textbf{Blend}, is designed for both speed and efficacy, hence targeting unlearning. Its primary innovation is a reparameterization of the distribution matching objective that significantly reduces the number of trainable parameters compared to SOTA like Distribution Matching (\textbf{DM}), Gradient Matching (\textbf{GM}) and Improved Distribution Matching (\textbf{IDM}) \cite{zhao2023dataset, zhao2020dataset, zhao2023improved}. For instance, on CIFAR-10 with an Images Per Class (IPC) of 10, \textbf{Blend} requires only $5 \times 10^4$ parameters, whereas DM requires $30.72 \times 10^4$.

We further shed light on this by conducting experiments on CNN and CIFAR-10. A key distinction in our setup is that \textbf{Blend}'s optimization process is limited to a synthetic data batch size of 1 in our implementation (as batched implementation is non-trivial). In contrast, competing methods like \textbf{GM}, \textbf{DM}, and \textbf{IDM} leverage large synthetic batch sizes (e.g., 256) for parallelized optimization. Despite being unable to use this common acceleration technique, \textbf{Blend}'s architectural efficiency leads to a dramatic performance gap. On CIFAR-10, \textbf{Blend} completes condensation in just 8-18 seconds (0.13–0.31 minutes). In stark contrast, GM, DM, and IDM require roughly 75–300 minutes, making our method over 1500x faster on average.

This dramatic speed-up does not compromise performance. As shown in Figure 2, \textbf{Blend} consistently yields higher test accuracy (e.g., 35.03\% at IPC=10) compared to \textbf{GM} (33.01\%), \textbf{DM} (32.2\%), and \textbf{IDM} (33.1\%). Furthermore, the synthetic prototypes generated by \textbf{Blend} are qualitatively superior. They are noticeably smoother and more class-consistent (e.g., clear airplane outlines) than the noisy, cluttered images from other methods. This demonstrates that our approach of grouping and blending feature-similar images is highly effective at producing coherent and informative exemplars.

{
\scriptsize                     
\renewcommand{\arraystretch}{0.8}   
\setlength{\tabcolsep}{4pt}         
\begin{table*}[t]
\centering
\caption{Unlearning ResNet-18/CIFAR-10 (mean $\pm$ 95\% CI) with $k$-means partitioning time $5.5753 \pm 0.1391$ s, $\mathcal{F}$-sampling time $0.1184 \pm 0.0107$ s}
\label{tab:resnet18_32_cifar100}
\resizebox{\textwidth}{!}{
\begin{tabular}{lcccccccccc}
\toprule
\multicolumn{1}{c}{} & 
\multicolumn{2}{c}{\textbf{MIA Score~$\downarrow$}} & 
\multicolumn{2}{c}{\textbf{Retain Accuracy~$\uparrow$}} & 
\multicolumn{2}{c}{\textbf{Forget Accuracy~$\downarrow$}} & 
\multicolumn{2}{c}{\textbf{Test Accuracy~$\uparrow$}} & 
\multicolumn{2}{c}{\textbf{Unlearning Time (s)~$\downarrow$}} \\
\cmidrule(lr){2-3} \cmidrule(lr){4-5} \cmidrule(lr){6-7} \cmidrule(lr){8-9} \cmidrule(lr){10-11}
\textbf{Method} & \textbf{w/o DC} & \textbf{w DC} & \textbf{w/o DC} & \textbf{w DC} & \textbf{w/o DC} & \textbf{w DC} & \textbf{w/o DC} & \textbf{w DC} & \textbf{w/o DC} & \textbf{w DC} \\
\midrule
\multicolumn{11}{c}{\textbf{Random Class Forgetting (Condensation Time for 1000 IPC: $47.7812 \pm 1.3323$ s leading to 80.0\% retain dataset reduction)}} \\
\midrule
Retraining       & $49.76 \pm 2.57$ & $50.51 \pm 3.55$ & \best{$100.00 \pm 0.00$} & $75.50 \pm 5.63$ & \best{$0.00 \pm 0.00$} & \best{$0.00 \pm 0.00$} & $80.35 \pm 2.08$ & $62.47 \pm 4.88$ & $1141.71 \pm 51.30$ & $224.50 \pm 0.14$ \\
CF               & $50.18 \pm 1.16$ & $55.39 \pm 2.67$ & $100.00 \pm 0.00$ & $96.39 \pm 0.75$ & $0.00 \pm 0.00$ & $0.30 \pm 0.45$ & \best{$81.19 \pm 2.40$} & \best{$77.75 \pm 1.25$} & $577.81 \pm 55.36$ & $112.32 \pm 0.04$ \\
Bad Distillation & $62.83 \pm 9.14$ & $57.31 \pm 7.01$ & $96.26 \pm 0.72$ & $98.88 \pm 0.18$ & $1.23 \pm 3.90$ & $24.99 \pm 79.51$ & $77.33 \pm 2.37$ & $81.19 \pm 6.28$ & $284.54 \pm 28.72$ & $117.04 \pm 0.05$ \\
$L_1$ Sparse     & \best{$47.32 \pm 2.73$} & $53.79 \pm 0.88$ & $99.00 \pm 0.97$ & $96.36 \pm 0.55$ & $0.00 \pm 0.00$ & $0.00 \pm 0.00$ & $78.42 \pm 3.40$ & $77.66 \pm 1.32$ & $358.62 \pm 28.80$ & $69.88 \pm 0.01$ \\
SCRUB            & $50.12 \pm 0.37$ & $50.00 \pm 0.00$ & $15.28 \pm 7.66$ & $11.11 \pm 0.00$ & $0.00 \pm 0.00$ & $0.00 \pm 0.00$ & $13.80 \pm 6.90$ & $10.00 \pm 0.00$ & $241.23 \pm 23.22$ & $61.58 \pm 0.02$ \\
Pruning          & $50.87 \pm 3.22$ & $50.80 \pm 2.27$ & $98.08 \pm 0.73$ & $100.00 \pm 0.00$ & $0.00 \pm 0.00$ & $0.00 \pm 0.00$ & $77.43 \pm 2.41$ & $72.65 \pm 1.27$ & $235.50 \pm 6.20$ & $46.74 \pm 0.01$ \\
A-CF  & $51.71 \pm 2.29$ &
\best{$48.25 \pm 7.31$} &
$97.88 \pm 4.39$ &
\best{$98.92 \pm 0.62$} &
$0.00 \pm 0.00$ &
$0.00 \pm 0.00$ &
$77.01 \pm 3.20$ &
$77.45 \pm 3.63$ &
\best{$75.76 \pm 0.27$} &
\best{$16.01 \pm 0.96$}
 \\
\midrule
\multicolumn{11}{c}{\textbf{10\% Uniform Forgetting (Condensation Time for 1000 IPC: $34.1780 \pm 0.7403$ s leading to 39.8\% retain dataset reduction)}} \\
\midrule
Retraining       & \best{$50.47 \pm 0.61$} & \best{$50.05 \pm 1.64$} & \best{$100.00 \pm 0.00$} & $91.48 \pm 0.41$ & $87.48 \pm 1.01$ & $82.25 \pm 1.01$ & $86.84 \pm 0.62$ & $80.65 \pm 0.48$ & $1132.22 \pm 33.71$ & $664.73 \pm 6.16$ \\
CF               & $52.45 \pm 1.18$ & $54.39 \pm 0.89$ & $100.00 \pm 0.00$ & $96.95 \pm 0.17$ & $93.30 \pm 0.53$ & $93.42 \pm 0.08$ & \best{$88.11 \pm 0.25$} & \best{$86.60 \pm 0.20$} & $561.76 \pm 4.92$ & $332.23 \pm 2.78$ \\
Bad Distillation & $82.92 \pm 1.51$ & $85.78 \pm 2.43$ & $93.78 \pm 0.64$ & $94.43 \pm 0.98$ & \best{$27.99 \pm 9.74$} & \best{$18.71 \pm 9.63$} & $78.77 \pm 0.56$ & $79.57 \pm 1.42$ & $275.64 \pm 0.56$ & $264.16 \pm 1.54$ \\
$L_1$ Sparse     & $50.87 \pm 1.39$ & $52.22 \pm 0.99$ & $98.87 \pm 0.23$ & $92.81 \pm 0.45$ & $86.18 \pm 0.60$ & $86.30 \pm 1.07$ & $84.36 \pm 0.35$ & $82.10 \pm 0.56$ & $348.20 \pm 0.98$ & $206.46 \pm 1.54$ \\
SCRUB            & $51.27 \pm 0.42$ & $51.59 \pm 1.46$ & $99.96 \pm 0.01$ & $77.94 \pm 42.33$ & $88.35 \pm 1.39$ & $73.11 \pm 37.70$ & $85.99 \pm 1.09$ & $70.56 \pm 34.47$ & $235.68 \pm 1.63$ & $149.79 \pm 0.95$ \\
Pruning          & $51.08 \pm 0.72$ & $51.75 \pm 0.68$ & $96.76 \pm 0.87$ & $97.52 \pm 1.28$ & $85.67 \pm 0.99$ & $83.22 \pm 2.08$ & $82.51 \pm 0.70$ & $79.75 \pm 2.08$ & $234.11 \pm 0.35$ & $138.16 \pm 0.96$ \\
A-CF  & $51.86 \pm 0.56$ &
$51.95 \pm 0.47$ &
$99.53 \pm 0.85$ &
\best{$99.76 \pm 0.10$} &
$86.19 \pm 1.41$ &
$85.42 \pm 0.15$ &
$86.18 \pm 2.50$ &
$86.23 \pm 0.38$ &
\best{$76.24 \pm 1.13$} &
\best{$47.32 \pm 3.05$}
\\
\bottomrule
\end{tabular}
}
\end{table*}
}

\subsection{Single-Round Unlearning}

We benchmark single-round unlearning across \textbf{utility} metrics (Test/Retain Accuracy; higher is better) and \textbf{privacy} metrics (Forget Accuracy/MIA Score; lower is better). We analyze two orthogonal improvements: (i) our \textbf{Blend} dataset condensation method, which shrinks the retain set, and (ii) our loss-centric \textbf{A-AMU} algorithm, only applied as acceleration of \textbf{CF}$\rightarrow$\textbf{A-CF} for brevity. The unlearning time in tables is separated from overhead from $k$-means partitioning and the Blend condensation (in case of “w/ DC”) for distinction.
\vspace{-1ex}
\vspace{-1ex}

\paragraph{Situational Impact of Blend}
The effectiveness of \textbf{Blend} scales with the data reduction ratio. For \textit{random class removal} on SVHN, shrinking the retain set to just 2.2\% enabled the \textbf{Pruning} method to achieve a 96.3\% total time reduction (from 647.4s to 23.7s) at the cost of only 3.9 percentage points in Test Accuracy, though the MIA score increased by 2.4. On CIFAR-10, a reduction to 20.0\% of the retain set gave \textbf{Retraining} a 75.7\% time reduction, but with a more substantial utility drop of 17.9 points in Test Accuracy and 24.5 points in Retain Accuracy. For \textit{10\% uniform forgetting} on CIFAR-10 (60.2\% retain), the gains were more modest: \textbf{Retraining} time was reduced by 37.8\%, with a 6.2-point drop in Test Accuracy and a slight improvement in the MIA score (from 50.47 to 50.05). Thus, \textbf{Blend} offers dramatic speed-ups when data reduction is large, but this can introduce a trade-off with model utility.
\vspace{-1ex}
\vspace{-1ex}

\paragraph{Consistent Acceleration with A-AMU}
Our \textbf{A-AMU} framework provides consistent algorithmic acceleration, independent of the unlearning scenario. On CIFAR-10 and without condensation, \textbf{A-CF} is \emph{7.6x} faster than naive CF for class removal (75.8s vs. 577.8s) and \emph{7.4x} faster for uniform forgetting (76.2s vs. 561.8s). This performance is achieved using a fixed set of hyperparameters for \textbf{A-CF} across all tasks.

{
\scriptsize                     
\renewcommand{\arraystretch}{0.8}   
\setlength{\tabcolsep}{4pt}         
\begin{table*}[t]
\centering
\caption{Unlearning on ResNet-18/SVHN (mean $\pm$ 95\% CI) with $k$-means partitioning time $4.1379 \pm 0.3221$ s, $\mathcal{F}$-sampling time $0.0537 \pm 0.0089$ s}
\label{tab:resnet18_32_cifar100_condensed}
\resizebox{\textwidth}{!}{
\begin{tabular}{lcccccccccc}
\toprule
\multicolumn{1}{c}{} & 
\multicolumn{2}{c}{\textbf{MIA Score~$\downarrow$}} & 
\multicolumn{2}{c}{\textbf{Retain Accuracy~$\uparrow$}} & 
\multicolumn{2}{c}{\textbf{Forget Accuracy~$\downarrow$}} & 
\multicolumn{2}{c}{\textbf{Test Accuracy~$\uparrow$}} & 
\multicolumn{2}{c}{\textbf{Unlearning Time (s)~$\downarrow$}} \\
\cmidrule(lr){2-3} \cmidrule(lr){4-5} \cmidrule(lr){6-7} \cmidrule(lr){8-9} \cmidrule(lr){10-11}
\textbf{Method} & \textbf{w/o DC} & \textbf{w DC} & \textbf{w/o DC} & \textbf{w DC} & \textbf{w/o DC} & \textbf{w DC} & \textbf{w/o DC} & \textbf{w DC} & \textbf{w/o DC} & \textbf{w DC} \\
\midrule
\multicolumn{11}{c}{\textbf{Random Class Forgetting (Condensation Time for 100 IPC: $4.90615 \pm 0.1941$ s leading to 97.8\% retain dataset reduction)}} \\
\midrule
Retraining & $50.45 \pm 1.53$ & $52.01 \pm 5.38$ & $99.99 \pm 0.00$ & $78.46 \pm 3.28$ & \best{$0.00 \pm 0.00$} & \best{$0.00 \pm 0.00$} & \best{$82.71 \pm 10.54$} & $68.53 \pm 8.79$ & $1012.11 \pm 4.18$ & $22.79 \pm 0.04$ \\
CF & \best{$50.34 \pm 3.97$} & \best{$51.15 \pm 0.32$} & \best{$100.00 \pm 0.00$} & $91.37 \pm 0.40$ & $0.00 \pm 0.00$ & $0.00 \pm 0.00$ & $82.37 \pm 10.30$ & $81.30 \pm 0.25$ & $505.95 \pm 2.05$ & $11.36 \pm 0.01$ \\
Bad Distillation & $55.38 \pm 1.74$ & $49.58 \pm 6.42$ & $98.49 \pm 2.02$ & $99.69 \pm 0.72$ & $0.00 \pm 0.00$ & $0.00 \pm 0.00$ & $81.84 \pm 9.47$ & $83.04 \pm 9.03$ & $248.95 \pm 1.34$ & $45.15 \pm 0.03$ \\
$L_1$ Sparse & $53.71 \pm 3.01$ & $52.07 \pm 1.39$ & $99.16 \pm 0.96$ & $96.97 \pm 0.13$ & $0.00 \pm 0.00$ & $0.00 \pm 0.00$ & $84.04 \pm 1.87$ & \best{$94.18 \pm 0.19$} & $314.88 \pm 3.28$ & $7.14 \pm 0.00$ \\
SCRUB & $50.47 \pm 1.64$ & $49.97 \pm 0.03$ & $12.93 \pm 2.81$ & $11.11 \pm 0.00$ & $0.00 \pm 0.00$ & $0.00 \pm 0.00$ & $13.61 \pm 8.37$ & $7.37 \pm 0.71$ & $211.82 \pm 2.00$ & $19.57 \pm 0.02$ \\
Pruning & $49.77 \pm 2.35$ & $51.90 \pm 2.20$ & $99.04 \pm 0.18$ & \best{$100.00 \pm 0.00$} & $0.00 \pm 0.00$ & $0.00 \pm 0.00$ & $80.80 \pm 10.02$ & $77.36 \pm 8.78$ & $210.93 \pm 1.00$ & $4.80 \pm 0.01$ \\
A-CF  & $52.04 \pm 2.70$ &
$54.09 \pm 6.30$ &
$99.65 \pm 0.59$ &
$97.55 \pm 0.71$ &
$0.00 \pm 0.00$ &
$0.00 \pm 0.00$ &
$80.90 \pm 9.38$ &
$84.49 \pm 2.80$ &
\best{$83.27 \pm 0.48$} &
\best{$2.07 \pm 0.02$}
\\

\midrule
\multicolumn{11}{c}{\textbf{10\% Uniform Forgetting (Condensation Time for 100 IPC: $0.4733 \pm 0.0759$ s leading to 1.6\% retain dataset reduction)}} \\
\midrule
Retraining & \best{$51.75 \pm 1.91$} & \best{$51.61 \pm 3.54$} & $99.99 \pm 0.00$ & $99.76 \pm 0.03$ & $94.60 \pm 0.59$ & $94.71 \pm 0.26$ & \best{$95.18 \pm 0.16$} & \best{$95.12 \pm 0.17$} & $1011.96 \pm 6.45$ & $992.98 \pm 7.97$ \\
CF & $53.67 \pm 0.51$ & $53.32 \pm 0.37$ & \best{$100.00 \pm 0.00$} & $99.86 \pm 0.02$ & $97.20 \pm 0.03$ & $97.06 \pm 0.55$ & $94.89 \pm 0.07$ & $95.09 \pm 0.22$ & $505.06 \pm 3.72$ & $496.01 \pm 3.47$ \\
Bad Distillation & $92.23 \pm 0.43$ & $93.57 \pm 0.44$ & $96.94 \pm 1.02$ & $99.75 \pm 0.11$ & \best{$23.77 \pm 8.50$} & \best{$23.77 \pm 2.00$} & $89.73 \pm 2.44$ & $91.72 \pm 0.41$ & $248.75 \pm 0.95$ & $369.65 \pm 2.02$ \\
$L_1$ Sparse & $53.55 \pm 1.96$ & $53.12 \pm 2.44$ & $99.32 \pm 0.44$ & $99.47 \pm 0.21$ & $93.80 \pm 1.00$ & $94.32 \pm 0.87$ & $93.60 \pm 0.57$ & $93.33 \pm 0.20$ & $314.80 \pm 2.90$ & $308.10 \pm 1.58$ \\
SCRUB & $54.99 \pm 1.47$ & $56.51 \pm 4.66$ & $99.66 \pm 0.07$ & \best{$99.99 \pm 0.02$} & $95.00 \pm 0.29$ & $99.07 \pm 0.44$ & $94.48 \pm 0.31$ & $95.13 \pm 0.19$ & $211.64 \pm 0.85$ & $214.29 \pm 1.76$ \\
Pruning & $54.20 \pm 1.60$ & $53.55 \pm 0.72$ & $99.33 \pm 0.10$ & $99.18 \pm 0.40$ & $94.91 \pm 0.16$ & $94.94 \pm 0.71$ & $93.34 \pm 0.41$ & $93.50 \pm 0.52$ & $210.79 \pm 1.00$ & $206.95 \pm 1.34$ \\
A-CF  & $54.89 \pm 2.23$ &
$53.96 \pm 1.56$ &
$99.48 \pm 0.47$ &
$99.48 \pm 0.32$ &
$93.13 \pm 1.94$ &
$93.50 \pm 1.46$ &
$94.09 \pm 1.06$ &
$93.66 \pm 1.56$ &
\best{$85.83 \pm 2.01$} &
\best{$82.83 \pm 1.35$}
\\

\bottomrule
\end{tabular}}
\end{table*}
}

{
\scriptsize                     
\renewcommand{\arraystretch}{0.8}   
\setlength{\tabcolsep}{4pt}         
\begin{table*}[t]
\centering
\caption{Unlearning on ResNet-50/SVHN (mean $\pm$ 95\% CI) with $k$-means partitioning time $4.1379 \pm 0.3221$ s, $\mathcal{F}$-sampling time $0.0537 \pm 0.0089$ s}
\label{tab:resnet50_32_cifar100_condensed}
\resizebox{\textwidth}{!}{
\begin{tabular}{lcccccccccc}
\toprule
\multicolumn{1}{c}{} & 
\multicolumn{2}{c}{\textbf{MIA Score~$\downarrow$}} & 
\multicolumn{2}{c}{\textbf{Retain Accuracy~$\uparrow$}} & 
\multicolumn{2}{c}{\textbf{Forget Accuracy~$\downarrow$}} & 
\multicolumn{2}{c}{\textbf{Test Accuracy~$\uparrow$}} & 
\multicolumn{2}{c}{\textbf{Unlearning Time (s)~$\downarrow$}} \\
\cmidrule(lr){2-3} \cmidrule(lr){4-5} \cmidrule(lr){6-7} \cmidrule(lr){8-9} \cmidrule(lr){10-11}
\textbf{Method} & \textbf{w/o DC} & \textbf{w DC} & \textbf{w/o DC} & \textbf{w DC} & \textbf{w/o DC} & \textbf{w DC} & \textbf{w/o DC} & \textbf{w DC} & \textbf{w/o DC} & \textbf{w DC} \\
\midrule
\multicolumn{11}{c}{\textbf{Random Class Forgetting} (\textbf{Condensation Time for 100 IPC: $4.90615 \pm 0.1941$ s leading to 97.8\% retain dataset reduction))}} \\
\midrule
Retraining & $51.57 \pm 0.76$ & $51.41 \pm 0.77$ & \best{$99.97 \pm 0.01$} & $40.85 \pm 16.59$ & \best{$0.00 \pm 0.00$} & \best{$0.00 \pm 0.00$} & $85.33 \pm 9.14$ & $36.53 \pm 16.44$ & $3177.10 \pm 18.08$ & $71.35 \pm 0.22$ \\
CF & $51.34 \pm 1.03$ & \best{$46.39 \pm 2.68$} & $99.97 \pm 0.02$ & $94.73 \pm 0.02$ & $0.00 \pm 0.00$ & $62.04 \pm 0.09$ & $85.45 \pm 9.50$ & \best{$88.79 \pm 0.56$} & $1587.86 \pm 8.52$ & $35.61 \pm 0.09$ \\
Bad Distillation & \best{$48.79 \pm 4.05$} & $49.77 \pm 8.64$ & $99.91 \pm 0.01$ & $99.82 \pm 0.38$ & $ 0.00 \pm 0.09$ & $00.00 \pm 0.87$ & \best{$89.99 \pm 5.84$} & $87.59 \pm 6.84$ & $1217.71 \pm 6.58$ & $221.71 \pm 0.58$ \\
$L_1$ Sparse & $52.67 \pm 5.92$ & $59.40 \pm 2.35$ & $99.24 \pm 0.10$ & $96.64 \pm 0.30$ & $0.00 \pm 0.00$ & $0.00 \pm 0.00 $ & $83.98 \pm 9.43$ & $83.07 \pm 1.00$ & $1298.78 \pm 6.70$ & $29.29 \pm 0.06$ \\
SCRUB & $49.99 \pm 0.04$ & $49.97 \pm 0.04$ & $11.70 \pm 0.70$ & $11.11 \pm 0.00$ & $0.00 \pm 0.00$ & $0.00 \pm 0.00$ & $12.08 \pm 9.45$ & $7.63 \pm 0.13$ & $679.75 \pm 3.76$ & $63.05 \pm 0.19$ \\
Pruning & $51.56 \pm 3.15$ & $53.94 \pm 3.66$ & $99.00 \pm 0.23$ & \best{$100.00 \pm 0.00$} & $0.00 \pm 0.00$ & $0.00 \pm 0.00$ & $84.31 \pm 9.53$ & $80.41 \pm 3.55$ & $647.44 \pm 4.07$ & $14.62 \pm 0.05$ \\
A-CF  & $48.98 \pm 3.16$ & $53.30 \pm 7.61$ & $97.31 \pm 0.84$ & $94.92 \pm 0.73$ & $0.00 \pm 0.00$ & $0.00 \pm 0.00$ & $85.92 \pm 3.27$ & $78.39 \pm 8.36$ & \best{$211.69 \pm 1.13$} & \best{$4.99 \pm 0.02$}
\\

\midrule
\multicolumn{11}{c}{\textbf{10\% Uniform Forgetting (Condensation Time for 100 IPC: $0.47335 \pm 0.0759$ s leading to 1.6\% retain dataset reduction)}} \\
\midrule
Retraining & \best{$51.68 \pm 4.66$} & \best{$51.03 \pm 3.50$} & \best{$99.98 \pm 0.02$} & $99.75 \pm 0.04$ & $94.62 \pm 0.14$ & $94.44 \pm 0.14$ & $95.00 \pm 0.15$ & $94.86 \pm 0.30$ & $3181.39 \pm 17.57$ & $3124.95 \pm 19.68$ \\
CF & $54.67 \pm 1.51$ & $53.69 \pm 2.54$ & $99.97 \pm 0.00$ & $99.87 \pm 0.03$ & $97.53 \pm 0.12$ & $97.48 \pm 0.07$ & \best{$95.22 \pm 0.13$} & \best{$95.24 \pm 0.11$} & $1587.19 \pm 9.69$ & $1561.58 \pm 12.57$ \\
Bad Distillation & $92.25 \pm 0.98$ & $92.76 \pm 1.05$ & $96.90 \pm 0.84$ & $99.71 \pm 0.20$ & \best{$20.13 \pm 9.61$} & \best{$22.32 \pm 4.01$} & $89.47 \pm 1.19$ & $92.16 \pm 1.11$ & $1218.23 \pm 7.08$ & $1818.03 \pm 11.68$ \\
$L_1$ Sparse & $54.98 \pm 4.00$ & $54.43 \pm 1.57$ & $99.16 \pm 0.21$ & $99.50 \pm 0.15$ & $94.06 \pm 0.97$ & $95.77 \pm 0.31$ & $93.13 \pm 1.06$ & $93.82 \pm 0.91$ & $1299.93 \pm 6.71$ & $1278.57 \pm 12.03$ \\
SCRUB & $53.62 \pm 1.89$ & $57.10 \pm 2.62$ & $99.17 \pm 0.63$ & \best{$99.96 \pm 0.04$} & $93.29 \pm 1.02$ & $99.06 \pm 0.35$ & $93.58 \pm 1.16$ & $95.17 \pm 0.31$ & $681.54 \pm 3.61$ & $690.14 \pm 5.75$ \\
Pruning & $54.42 \pm 0.62$ & $55.28 \pm 4.82$ & $99.06 \pm 0.21$ & $98.39 \pm 1.19$ & $95.03 \pm 0.67$ & $94.17 \pm 1.87$ & $93.73 \pm 0.37$ & $91.82 \pm 3.67$ & $647.93 \pm 3.91$ & $637.52 \pm 5.97$ \\
A-CF  & $54.89 \pm 1.89$ &
$54.29 \pm 2.43$ &
$98.35 \pm 1.00$ &
$96.82 \pm 1.24$ &
$93.10 \pm 1.92$ &
$91.46 \pm 1.00$ &
$93.41 \pm 1.64$ &
$92.24 \pm 1.24$ &
\best{$214.15 \pm 2.44$} &
\best{$207.61 \pm 0.83$}
\\
\bottomrule
\end{tabular}}
\end{table*}
}

\begin{figure*}[b]
\centering
\includegraphics[height=0.14\linewidth, width=1.0\linewidth]{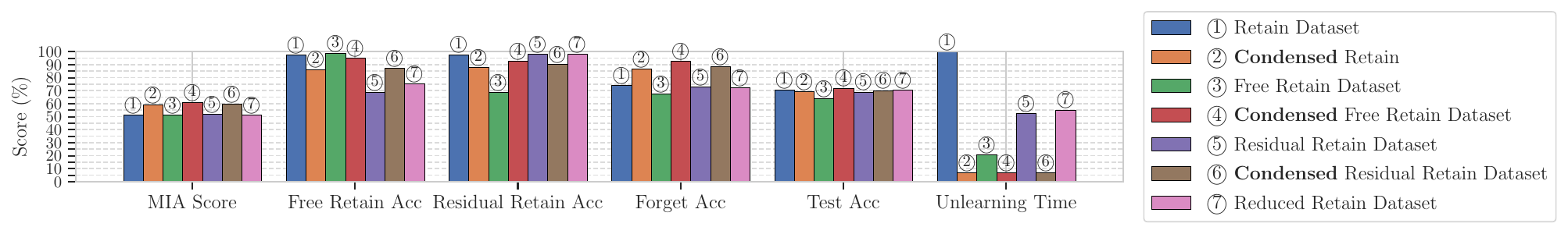}
\caption{Impact of different condensation strategies (on \textbf{Blend}) on 10\% random unlearning performance of $\mathbf{L_1}$-sparse on ResNet-18/CINIC-10}
\label{fig:short}
\end{figure*}

\begin{figure*}[t]
\centering
\includegraphics[height=0.19\linewidth, width=1.0\linewidth]{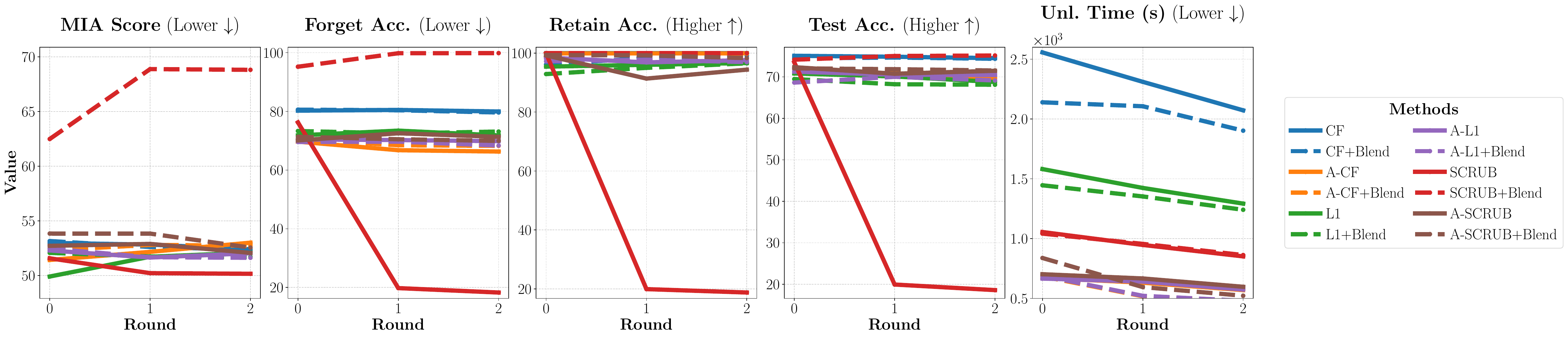}
\caption{Multi-round sequential unlearning (random 10 \%) performance of without accelerated \textbf{CF}, \textbf{L1}-sparse, and \textbf{SCRUB}, with acceleration (\textbf{A-CF}, $\textbf{A-}\mathbf{L_1}$, \textbf{A-SCRUB}), with and without dataset condensation (\textbf{Blend}). With dataset condensation allows 34.9\% retain dataset reduction}
\label{fig:short}
\end{figure*}

\vspace{-1ex}
\vspace{-1ex}

\paragraph{Combined Impact}
When our two proposals are combined, the outcome depends on the scenario. For class removal, adding \textbf{Blend} to the already-fast \textbf{A-CF} provides a further 8.4\% speed-up (from 75.8s to 69.4s total time). However, for uniform forgetting, where data reduction is modest, \textbf{Blend}'s overhead leads to a 14.2\% slowdown (from 76.2s to 87.1s). This indicates that for fast algorithms with limited data reduction, standalone \textbf{A-AMU} is the superior strategy.

\vspace{-1ex}
\vspace{-1ex}

\paragraph{Bird’s-Eye View}
In summary, our proposals offer two distinct tools. \textbf{Blend} is a powerful situational method whose time savings are proportional to the data reduction, though this may involve a trade-off with model utility. \textbf{A-CF} is a universal, tuning-free algorithmic accelerator. Ultimately, \textbf{A-CF} delivers the best overall balance of speed, utility, and privacy, consistently performing within 5 percentage points of the best-in-class score across every setting we tested, making it the most robust and practical solution.

\subsection{Empirical Justification of Reduced Retain
Dataset as Condensed Dataset}
We evaluate the impact of various condensation strategies on the retain set for the $\mathbf{L_1}$-sparsity unlearning method (Figure~3). Experiments use a ResNet-18 pretrained on CINIC-10 with 10\% uniform unlearning. The model's extreme overfitting (99.98\% train vs.\ 73.16\% test accuracy) thus represents a worst-case, poor-generalization scenario to stress-test condensation effects \cite{li2025mubox}, as iterative unlearning fails when gradients vanish on overfitted data. Building upon Section 3.3, we systematically quantify how naive condensation trades off privacy and efficiency under these challenging conditions.

\begin{itemize}
  \item \textbf{Full condensation dulls unlearning (\circled{2}).}  
        Forget-accuracy rises to $\approx87\%$ and the MIA score
        increases, confirming that global prototypes produce features arbitrarily close to features of $\mathbf{F}$.
  \item \textbf{Single-partition training is lopsided
        (\circled{3}: raw $\mathbf{R_{\text{free}}}$, \circled{5}: raw $\mathbf{R_{\text{residual}}}$).}  
        Training only on $\mathbf{R_{\text{free}}}$ lowers forget-accuracy to
        $\approx67\%$ but also drags test accuracy on the residual side
        to $\approx64\%$.  
        Training only on $\mathbf{R_{\text{residual}}}$ keeps generalization
        high, yet leaves forget-accuracy essentially unchanged,
        revealing a privacy-utility trade-off.
  \item \textbf{Condensing the partition still dulls unlearning
        (\circled{4}: condensed-free, \circled{6}: condensed-residual).}  
        Summarizing $\mathbf{R_{\text{free}}}$ (\circled{4}) floods training with
        coarse prototypes that arbitrarily fall near forget features; and the model re-memorizes $\mathbf{F}$
        (forget-accuracy $\approx93\%$).  
        Summarizing $\mathbf{R_{\text{residual}}}$ (\circled{6}) removes boundary
        detail, pushing forget-accuracy back up to $\approx88\%$ and
        depressing \textit{free-retain} accuracy into the high-80s.
  \item \textbf{Selective condensation bridges every gap (\circled{7}).}  
        By \emph{preserving} boundary-critical residual images and
        \emph{condensing} only the distant free clusters, our reduced
        retain set  
        (i) reaches a forget-accuracy of $72.5\%$, matching the best
        single-partition strategy (\circled{5}) while dramatically
        outperforming the free-only alternative (\circled{3});  
        (ii) lifts \textit{free-retain} accuracy from the low-60s
        (\circled{5}) up to 75\%, closing the generalization gap
        without sacrificing privacy;  
        (iii) maintains the baseline MIA score; and  
        (iv) trims unlearning time from 100\% (\circled{1}) down to
        63.8\%, matching the speed of
        training on $\mathbf{R_{\text{residual}}}$, yet without its
        drop in free-retain performance.
\end{itemize}

\subsection{Multi-Round Unlearning}

We tested three base methods---\textbf{CF}, \textbf{L$_1$}-Sparsity, and \textbf{SCRUB}---and their accelerated variants---\textbf{A-CF}, $\textbf{A-}\mathbf{L_1}$, \textbf{A-SCRUB}) over three sequential 10\% uniform-forgetting rounds on ResNet-18 with CINIC-10. Experiments were run with and without \textbf{Blend} condensation, which adds a one-time 95.6s overhead. Figure 4 plots these results.

\vspace{-1ex}
\vspace{-1ex}

\paragraph{Distinct Roles of Blend and A-AMU}
Our two proposals address different challenges in sequential unlearning. \textbf{Blend} condensation acts as a stabilizer for unstable unlearning methods. This effect was visible on \textbf{SCRUB}, where it completely prevented the Test Accuracy collapse seen in the standalone version (which plummeted from 73.5\% to 18.6\%), maintaining performance at a stable $\approx75\%$ across all rounds. However, this utility stabilization came at the cost of privacy, as \textbf{SCRUB}'s Forget Accuracy and MIA score were notably higher with condensation (e.g., in Round 0, MIA score increased from 51.58\% to 62.48\%). For already stable methods, \textbf{Blend}'s impact on speed was modest (a 4-8\% time reduction for \textbf{CF}) to negligible (\textbf{L$_1$}-Sparsity). In contrast, our \textbf{A-AMU} framework provides a more substantial, upfront acceleration. In Round 0, standalone \textbf{A-CF} is \emph{3.8x faster} than \textbf{CF} (671.5s vs. 2556.6s), and \textbf{A-SCRUB} is \emph{1.5x faster} than \textbf{SCRUB} while also inherently preventing its performance collapse.

\vspace{-1ex}
\vspace{-1ex}

\paragraph{Combined Performance and Long-Term Benefits}
When both proposals are combined in this low-data-reduction scenario, \textbf{Blend}'s overhead leads to a slight initial slowdown for the already-fast accelerated methods. For instance, the total time for \textbf{A-CF} with condensation in Round 0 is 779.2s, compared to 671.5s for standalone \textbf{A-CF}. However, the combination shows a steeper decline in runtime over subsequent rounds, with the condensed version's runtime decreasing by 30.2\% from Round 0 to Round 2.

\vspace{-1ex}
\vspace{-1ex}

\paragraph{Bird's-Eye View}
Our multi-round analysis reveals two main findings. First, \textbf{Blend} condensation is a powerful tool for ensuring utility stability, rescuing volatile methods like \textbf{SCRUB} from catastrophic failure, albeit with a trade-off in privacy metrics. Second, our \textbf{A-AMU} framework provides a more significant and immediate acceleration (1.5-3.8x) than condensation in this high-frequency unlearning scenario. Together, they form a robust toolkit for sequential unlearning.

\vspace{-1ex}
\section{Conclusion}

In this work, we addressed the critical efficiency bottlenecks in approximate machine unlearning (AMU) by introducing a dual data- and loss-centric framework. ur first contribution, Blend, is a highly efficient dataset condensation method that dramatically reduces unlearning time in high-data-reduction scenarios like class removal. Our second, A-AMU, is a novel loss-centric framework that provides consistent, significant algorithmic acceleration and unparalleled hyperparameter stability. These proposals allow cutting end-to-end unlearning latency by 84.6\% in single-round and 54.5\% in multi-round settings. Loss-centric proposal consistently maintains a strong balance across all utility and privacy metrics, while data-centric see good gains when original unlearning time is large, and reduction of dataset is around 50\%. More broadly, we revitalize dataset condensation and MIA-based regularization as powerful tools for this domain. We hope our work paves the way for a new line of research focused on creating truly practical and efficient machine unlearning systems.

{\small
\bibliographystyle{ieee_fullname}
\bibliography{egbib}
}

\end{document}